\documentclass{article}
\usepackage{fullpage}
\usepackage{setspace}
\usepackage{amsmath,amssymb}
\usepackage{natbib}
\usepackage{algorithmic}
\usepackage[pdftex]{graphicx}
\usepackage[usenames,dvipsnames]{color}

\title{The Statistical Inefficiency of Sparse Coding for Images\\(or, One Gabor to Rule them All)}

\author{
James Bergstra,
Aaron Courville, and
Yoshua Bengio
}

\begin{document}

\maketitle

\begin{abstract}

Sparse coding is a proven principle for learning compact representations of images.  However, sparse coding by itself often leads to very redundant dictionaries. With images, this often takes the form of similar edge detectors which are replicated many times at various positions, scales and orientations. An immediate consequence of this observation is that the estimation of the dictionary components is not statistically efficient.  We propose a factored model in which factors of variation (e.g. position, scale and orientation) are untangled from the underlying Gabor-like filters.  There is so much redundancy in sparse codes for natural images that our model requires only a single dictionary element (a Gabor-like edge detector) to outperform standard sparse coding.  Our model scales naturally to arbitrary-sized images while achieving much greater statistical efficiency during learning. We validate this claim with a number of experiments showing, in part, superior compression of out-of-sample data using a sparse coding dictionary learned with only a single image.
\end{abstract}

\section{Introduction}

Sparse coding has been advanced as an inductive principle to build
efficient bases for pattern classification~\citep{RainaR2007} and to explain
the functional properties of cells in visual cortex and in auditory
cortex~\citep{Olshausen-97,Lewicki+Sejnowski-2000}.
But sparse coding is a general principle: there is nothing in the sparse coding algorithm that makes reference to images.

Sparse coding leads to compact representations at the expense of large overcomplete basis sets.  
Models of natural images require large dictionaries.
Large overcomplete image models require comensurate quantities of data to learn.
For example, the classical sparse coding result of \citet{Olshausen+Field-1996}
shows small filters (e.g. 12x12 pixels) whose training required tens of thousands of images.
Looking at the learned bases, it is clear that there is structure in the bases that the model has learned.
The vast majority of bases are localized edge detectors, placed in many locations, at many scales, and orientations.
By incorporating a prior that this structure should exist,
we show here that we can learn sparse coding dictionaries from much less data.

This may seem like an obvious point, but as it has become quite common in
the machine learning literature to use sparse coding as a means of learning
sparse representation spaces
\citep{RainaR2007,Bradley+Bagnell-2009,EladAharon2006}.
We believe it is
  appropriate to carefully evaluate the cost (in statistical efficiency)
  paid by not taking advantage of known transformational invariances.
  Further, statistical efficiency is necessary if we would like to apply
  sparse coding to full-size images.

Previous work such
  as~\citet{Grimes+Rao:2005,Grosse-2007} have presented strategies for
  convolutional sparse coding.  In these approaches, the overcomplete basis
  dictionary is essentially replicated at all locations.
Replication of a basis dictionary across all image locations is appropriate if we believe that the image statistics of interest are invariant across image locations.
Taking this line of thinking further, 
we can observe that the image statistics of interest for learning Gabor filters appear to be
invariant in scale and orientation.
In fact, once all of these invariances have been taken into account, 
there is really only one {\em kind} of dictionary element -- an oriented Gabor edge.

Bilinear modeling~\citep{tenenbaum00separating,Grimes+Rao:2005},
\begin{equation}
    \mathbf{z} \sim {\cal N} \left( \sum_{i=1}^{m}\sum_{j=1}^{n} \mathbf{w}_{ij} x_i y_j, \sigma^2 \right),
\end{equation}
has been put forward as a model for separating content from style.
Basis vectors $\mathbf{w}_{ij}$ are modulated by two vectors $\mathbf{x} \in \mathbb{R}^m$ and 
$\mathbf{y} \in \mathbb{R}^n$ of coefficients.
Bilinear models can separate content from style by using a Cartesian product of all possible sorts of content (ranged over by $i$ for example),
and all sorts of style (ranged over by $j$).
However, we would like for inference to recover both pieces of information:
what basis elements were most active, and how were they expressed?  
Bilinear modeling does not recover this information.
To see why -- consider that inference in this model recovers only $m+n$ coefficients, 
whereas there are $mn$ possible expressions of elements.
Factored 3-way Restricted Boltzmann Machines~\citep{ranzato+krizhevsky+hinton:2010}
have also been proposed as a means of separating content 
from style in the context of energy-based modeling, but they
share the same limitation.

This paper presents a variant of sparse coding in which the overcomplete basis
is factored into a set of basis vectors (``what'') and an infinite set of possible transformations
(``how'') that span a range of translations, rotations, and scales.
The set of transformations is rich enough that
our model can do effective sparse coding of images with just a single basis vector (which
unsurprisinly looks like an oriented Gabor edge).




\section{Factored Sparse Coding}

Our factored model can be developed from the basic linear generative model used in sparse coding:

\begin{equation}
    \mathbf{z} \sim {\cal N} \left( \sum_{i=1}^{m} \mathbf{w}_i x_i = W \mathbf{x}, \sigma^2 \right),
\end{equation}

in which $\mathbf{z}$ is a $k$-dimensional observation (in our case we will deal with images of $k$ pixels),
$\mathbf{w}_i$ is a $k$-dimensional basis vector, 
and $x_i$ its scalar coefficient in the expansion that generates $\mathbf{z}$.
We call $W$ a dictionary or code book because of its role in compression, where we transmit only the ($i,x_i$) pairs for which $|x_i|$ is largest, and the receiver must look
up these $i^{th}$ basis vectors to reconstruct the signal.

Our model goes further along the progression of introducing more factors of variation
and towards allowing more freedom in how they can be combined. In this paper, to
keep concepts simple and interpretable, 
we fix five dimensions of image variability:  horizontal scaling ($\alpha$), vertical scaling ($\beta$), rotation ($\theta$), horizontal translation ($\delta$), and vertical translation ($\eta$).
We also replace finite summations by integrals because the factors of variation are continuous quantities.
In our experiments we found that the transformations offered enough flexibility that we could set $n=1$ and drop the original sparse coding summation from the model entirely.
Our model then, is
\begin{equation}
    \mathbf{z} = 
        \int_{\alpha}^{} 
        \int_{\beta}^{} 
        \int_{\theta}^{} 
        \int_{\delta}^{} 
        \int_{\eta}^{} 
        \mathbf{w}_{\alpha\beta\theta\delta\eta} y_{\alpha\beta\theta\delta\eta}
        = \int_{\omega} \mathbf{w}_\omega y_\omega,
        \label{eq:fsc}
\end{equation}

with $w_\omega = \text{translate}(\text{rotate}(\text{scale}(\mathbf{u}, \alpha, \beta), \theta), \delta, \eta)$,
and $\mathbf{u}$ is a single generic filter (e.g. the generic Gabor filter).
Since the indices ($\omega=\{ \alpha,\beta,\theta,\delta,\eta \}$) are real, continuous values it can be helpful to think of $\mathbf{w}$ and $\mathbf{y}$ as real-valued functions, but we will continue to use a tensor index-notation to emphasize the connection with previous work.

We parametrize the dictionary $W$ by the sum of $m$ Dirac functions,
so in effect there are still $m$ points of support for $y_\omega$, and it makes sense to talk about a dictionary of size $m$.
The $m$ points of support are drawn i.i.d. from a prior on $\omega$, the details of the prior we used are given below in Section~\ref{sec:emp}.

Given the finite support of $m$ samples $\omega_i$, and $\mathbf{u}$,
inference of $\mathbf{x}^*$ for some observed $\mathbf{z}$ is just like sparse coding.
Learning in the model is also like in sparse coding, 
in the sense that for an $\mathbf{z},\mathbf{x}^*$ pair,
we take step in the direction of the negative gradient on $\mathbf{u}$.
In the factored model, 
we must apply the chain rule and back-propagate
reconstruction error through the translation, rotation, and scaling operators
and sum the gradient through each of the $m$ $\mathbf{w}$ terms.
All of these computations are $O(m)$ complexity in time and space.

Given this linear generative model, the idea of sparse coding is to encode
$\mathbf{z}$ by a vector $\mathbf{x}$ with a small number of non-zero
elements.  When $m \leq k$ sparse coding is not possible, but in an
overcomplete setting where $m \gg k$ there are many ways to encode
$\mathbf{z}$ and sparse coding typically selects one by the following likelihood maximization:

\begin{equation}
    \mathbf{x^*} = \arg \max_{\mathbf{x}} -\left| \left| \frac{\mathbf{z} - W \mathbf{x}}{\epsilon} \right| \right|^2 + \log P(\mathbf{x}),
\end{equation}

where $P(\mathbf{x})$ is a sparsity-inducing prior.
Several sparsity objectives (L1, Cauchy) and minimization techniques 
(conjugate gradient, linear programming) have been proposed~\citep{Olshausen-97,Lewicki+Sejnowski-2000,HonglakLee-2007}.
In our work, we take the Cauchy prior 
$P(x) \propto \prod_i e^{\gamma \log(1+\frac{x_i}{r^2})}$
used in~\citet{Olshausen-97}.
This prior is flat near $x_i=0$ so inference generally does not result in perfect zeros,
but it allows us to do inference by gradient descent, which is simple to implement.

{\bf Convexity}

One appeal of sparse coding is that inference (of $\mathbf{x}^*$ given $\mathbf{z}$ and $W$)
is convex. Interestingly, the factored sparse coding approach presented here retains this
property. Learning the dictionary (when coefficients $\mathbf{x}$ are not given) is
not convex for sparse coding and it is not here either. As we find in the
experiments below, compared to sparse coding,
inference and training are both much faster with the proposed model.

\section{Image Transformations}

This section describes how the generic filter $\mathbf{u}$ is transformed in our model to produce the dictionary $W$.

Recall that a transformation is specified by a tuple $\omega=\{ \alpha,\beta,\theta,\delta,\eta \}$.
This means to scale vertically by $e^\alpha$, scale horizontally by $e^\beta$, rotate by $\theta$,
translate vertically by $\delta$ and translate horizontally by $\eta$.

Mathematically, these transformations correspond to 3x3 linear transformations of 
 an input pixel coordinate $(u,v)$ at row $u$ and column $v$.
The transformation tells us, for any input location, at what output location $(r,c)$ should it be drawn.
\begin{align}
\begin{pmatrix} r\\ c \\1 \end{pmatrix}
    &= 
\begin{pmatrix} 1 & 0 & \delta\\ 0 & 1 &\eta \\ 0 & 0 & 1 \end{pmatrix}
\begin{pmatrix} \cos(\theta) & -\sin(\theta)&0\\ \sin(\theta)&\cos(\theta) &0 \\ 0 & 0 & 1 \end{pmatrix}
    \begin{pmatrix} e^{\alpha} &0&0\\ 0&e^{\beta} &0 \\ 0 & 0 & 1 \end{pmatrix}
\begin{pmatrix} u\\ v \\1 \end{pmatrix}
\end{align}

We define $(u,v)=0,0$ to be the middle of the input image and $(r,c)=(0,0)$ to be the middle of the output image.
Inconveniently, for integer-valued pixel input coordinates $u,v$, the output coordinates $r,c$ are typically not integer-valued and we must interpolate somehow.  But interpolation is non-trivial because the set of $r,c$ outputs that correspond to the set of $u,v$ inputs are not evenly spaced (when the image is scaled up for example, points near the origin are closer than points far from the origin).

Instead, we generate images by iterating over the output pixel locations.
For each output location $(r,c)$ 
we multiply by the inverse of the matrix product above to obtain 
a [generally] real-valued input pixel location $(u,v)$
    and set the output pixel value at position $(r,c)$ by bilinear interpolation in the input image nearest to position $(u,v)$.
    When multiplication by the inverse matrix yields a pixel location that is not on the image,
    we consider that all undefined pixels are zero.

\section{Empirical Results}\label{sec:emp}

We evaluated the factored sparse coding model with the CIFAR-10 dataset~\citep{KrizhevskyHinton2009}.
CIFAR-10 is a set of sixty thousand (60000) small color images of resolution 32x32.
The dataset images are downsampled images of everyday objects from Google image search, 
and they are manually labeled (selected) according to the main object in each image.
There are ten classes in the dataset: airplane, automobile, bird, cat, deer, dog, frog, horse, ship, truck.
In a first evaluation, in which we look at filters and data compression, the labels were not used.
In the second evaluation we look at the inferred codes as features for classification.
For all the experiments, the images are first whitened as in~\citet{Olshausen-97}.

\subsection{Dictionary Learning}\label{sec:empirical_learning}

\begin{figure}
\begin{center}
    (a)\includegraphics[width=.41\linewidth]{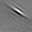}
    ~~~~~~~~~~~~(b)\includegraphics[width=.41\linewidth]{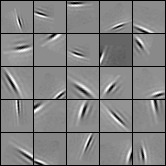}

\vspace*{8mm}

    ~(c)\includegraphics[width=.41\linewidth]{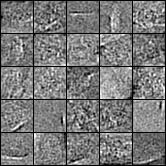}
    ~~~~~~~~~~~~(d)\includegraphics[width=.41\linewidth]{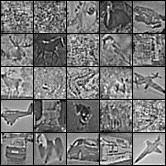}
\end{center}
\caption{(a) The lone generic filter $\mathbf{u}$ and (b) a subset of the dictionary $W$, learned by the factored sparse coding algorithm from the first 5000 images of the CIFAR-10 dataset ($m=1000$).
Conventional sparse coding dictionaries with 500 elements (c) and 4000 elements (d) are not fit well from so little data -- smaller dictionaries are no longer overcomplete, and larger dictionaries simply memorize training examples.
\label{fig:fsc_dict}
}
\end{figure}

The result of learning a factored sparse coding dictionary from the first 5000 CIFAR-10 images is illustrated in Figure~\ref{fig:fsc_dict}
CIFAR-10 images have 32x32 resolution, so we chose the following set $\Omega$:
$\alpha \in (-.4,.5)$,
$\beta \in (-.4,.5)$,
$\theta \in (0,2\pi)$,
$\alpha \in (-15,15)$
$\beta \in (-15,15)$.
Learning converged after between one and two passes through the data.
Minibatches of size 100 were used, 
and gradient updates to $\mathbf{u}$ were scaled by a learning rate of $2\times 10^{-5}$.
Inference of the sparse code was done with L-BFGS gradient descent {\tt scipy.optimize.fmin\_l\_bfgs\_b}, with the {\tt maxfun} set to 200.
Both $\mathbf{u}$ and every basis vector $\mathbf{w}_{\omega_i}$ were normalized to have unit length.
The dictionary with $m=500$ learned from random initial conditions in about 10 minutes on an NVidia GTX-285 GPU.

For purposes of comparison, we also learned conventional sparse coding dictionaries of a few sizes (500, 1000, 2000, 4000) on the same data.
Some of the dictionary elements from these dictionaries are illustrated in panels (c) and (d) of Figure~\ref{fig:fsc_dict}.
The sparse coding models took much longer to train, on the order of a few hours even on a GPU,
visibly poorer solutions (including noisy filters, and filters that replicate individual training examples).
The examples in CIFAR-10 have 1024 pixels, although there are far fewer important principle components,
so the dictionary of 500 units was bordering the non-overcomplete regime, however the dictionary of 2000 units contained many filters
which were simply copies of entire training examples.
There were not sufficient data to learn a good,
general overcomplete representation,
and this insufficiency of data is inevitable when basic sparse coding is applied to larger images.
{\bf Factorization of the dictionary makes learning statistically efficient.}

\subsection{Compression}\label{sec:compression}

\begin{figure}
        \centering
        \includegraphics[width=.46\linewidth]{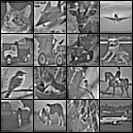} \hspace*{5mm}
        \includegraphics[width=.46\linewidth]{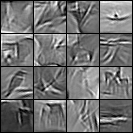}
 
        \hspace*{-1cm}(a) \hspace*{6cm}(b)

        \centering
        \includegraphics[width=0.9\linewidth]{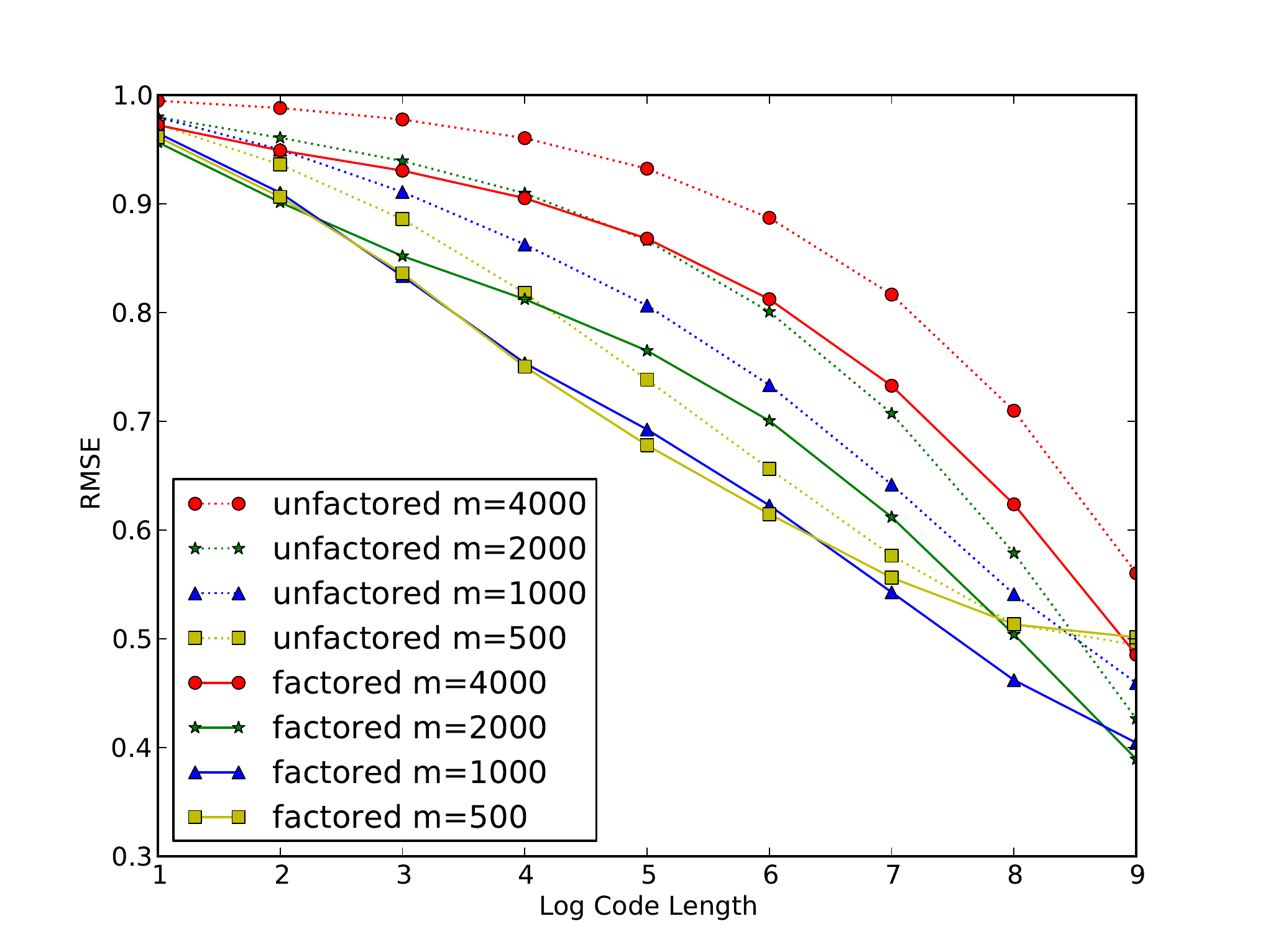}

    \vspace*{-3mm}\hspace*{5mm}(c)
    \caption{
    In panel (a), the factored coding model with 500 basis vectors reconstructs test images (b) 
    using only the largest 64 coefficients.
    Individual Gabor filters are still visible as they are combined together to form the reconstruction.
    Panel (c) shows the mean reconstruction error using only the largest $K$ coefficients 
    for each of the first 1000 test images from the CIFAR-10 dataset.
    For a variety of dictionary sizes ($m$) and code lengths,
    the factored sparse coding models (solid lines) 
    consistently require about half as many non-zero coefficients in order to
    produce the same reconstruction error as non-factored models (dashed lines).
    Estimation error on the RMSE values is negligeable, error bars have been omitted for clarity.
    (Lower is better.)
    }
    \label{fig:rmse}
\end{figure}

The factored sparse coding model is more statistically efficient than conventional sparse coding, when appropriate transformations are known.
To demonstrate this, we compared the coding efficiency of factored sparse coding versus conventional sparse coding using dictionaries learned from the first 5000 examples from CIFAR-10.
We ran the comparison at various amounts of capacity $m\in \{500, 1000, 2000, 4000\}$.
Subsets of the dictionaries involved are shown in Figure~\ref{fig:fsc_dict} in panels (b), (c), and (d).
Each dictionary was used to code out-of-sample data from the remaining examples of CIFAR-10.
We measured coding efficiency by the RMSE of the reconstructed image when retaining only the $K$ largest coding coefficients.
The performance of each model according to this measure is shown in Figure~\ref{fig:rmse}.

The factored models consistently outperformed the non-factored models.
At many points (capacity $m$ and number of non-zero coefficients $K$)
the factored model was able to match the RMSE of the non-factored model using twice as many coefficients,
corresponding to a {\em halving} of the expected code length.

It is interesting to note that the curves for the smaller models in Figure~\ref{fig:rmse}, the RMSE drops roughly linearly with the log-code-length, while for the larger models the drop is smaller for small code lengths and larger for larger code lengths.  This indicates that the first few dictionary elements chosen in large models are not as effective at coding the entire image.
Although the same effect is observed in the factored and un-factored models, we hypothesize different explanations in the two cases.
In the case of the unfactored model, we hypothesize that it is simply a result of overfitting of the model.
Looking at the dictionary (Figure~\ref{fig:fsc_dict}, panel (d)) suggests that many of the basis elements are simply copies of training examples. This makes a poor basis for coding test images, and so the RMSE curve of the unfactored model with $m=4000$ is higher (worse) than the rest.

In the case of the unfactored model, we postulate that it is a result of an inaccurate prior over transformations.  The large factored model learns a somewhat smaller Gabor edge than the smaller factored models, so that it can model finer detail when replicated all over it is replicated in so many locations that it can capture finer detail.
None of the models were optimized for a setting in which only a few largest coefficients would be used,
and the larger model is hurt most by this restriction. 

\begin{figure}
        \centering
            \includegraphics[width=.65\linewidth]{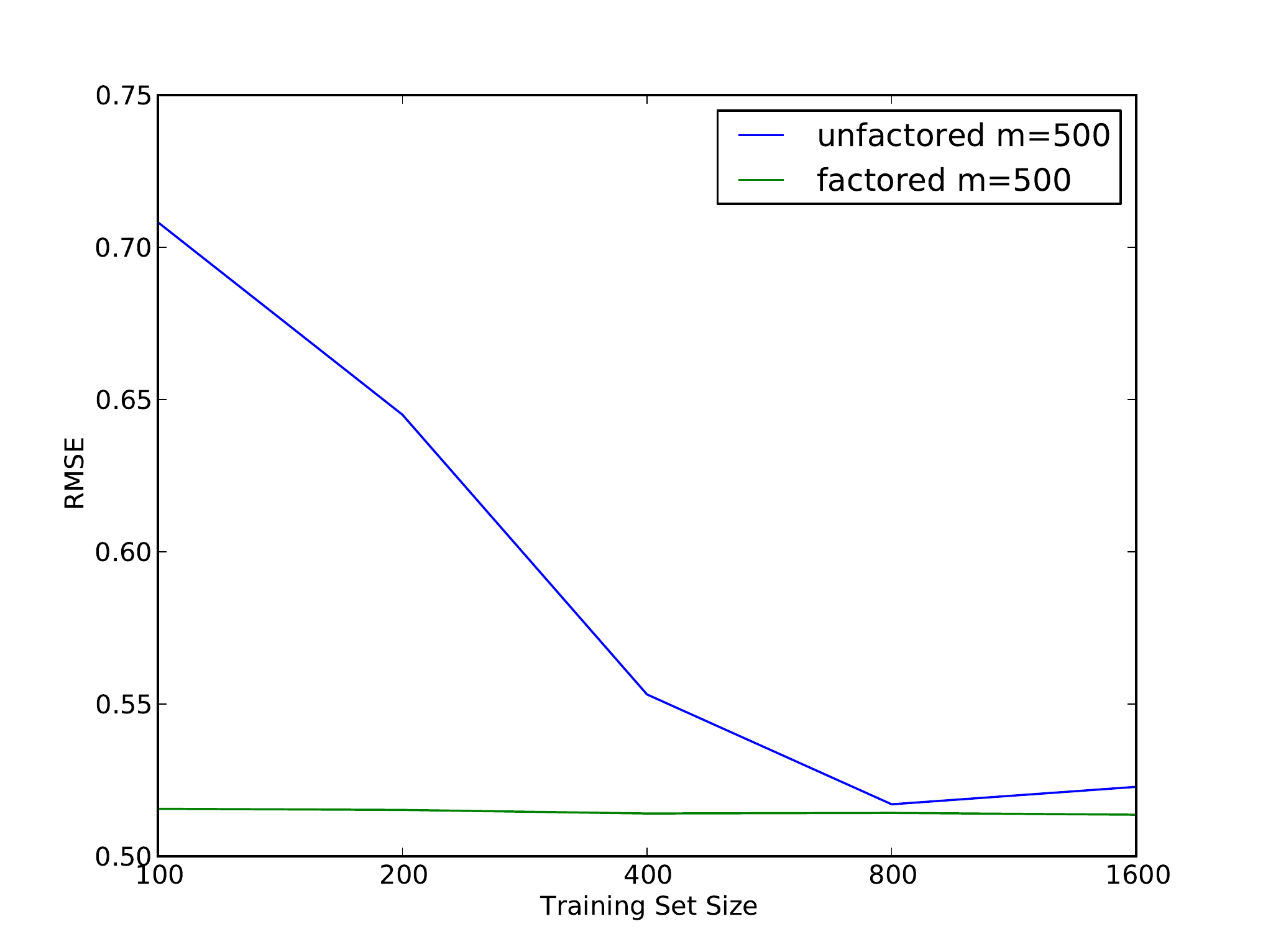}
    \caption{
    Unfactored sparse coding and factored sparse coding on out-of-sample data, using (all elements of) dictionaries learned from increasing amounts of data.
    Just 100 images is more than enough for the factored model to learn an idealized Gabor edge-filter, but a conventional sparse coding approach requires
    enough data to learn a Gabor in many locations.
    Compared with Figure~\ref{fig:rmse}, these values correspond to the right-most values, where $K=m$.
    \label{fig:lowdata}
    }
\end{figure}

Another way to visualize the data requirement of sparse dictionary learning is by looking at out-of-sample RMSE for models trained on various amounts of training data.
Figure~\ref{fig:lowdata} illustrates that for small amounts of data (even smaller than the dictionary size, and much smaller than the dimensionality of the signal),
the factored sparse coding model can learn as well as in the case of much more data.

\section{Conclusion}

We have set out to demonstrate that the common approach to sparse coding of
learning a highly redundant dictionary is statistically inefficient. As
image and video data continue to increase in resolution, practitioners ignore this
issue at their peril. We also take steps along a line of research (that
includes the work of \citet{Grimes+Rao:2005,Grosse-2007}) that deals
directly with the redundancy of sparse coding by factoring the sparse
coding model into a set of transformations and an invariant dictionary. In
our experiments the sparse coding dictionary was so redundant that we were
able to exceed its performance with a factored dictionary with only a
single invariance dictionary component.

In the factored model we consider, the transformations are fully specified
in advance. An obvious extention would have these transformations learned
as part of the dictionary learning scheme. This would be particularly
useful and interesting in moving beyond image applications where the set of
relevant transformations is relatively well understood.  In future work, we
also intend to pursue building hierarchical models of factored sparse
codes. We believe that our explicit representation of the basis
transformations makes our approach particularly amenable to learning higher
level structures in images and data.

\subsubsection*{Acknowledgments}

This work was supported by the National Science and Engineering Research Council of Canada (NSERC) and
the University of Montreal.


\bibliography{local,strings,ml,aigaion}
\bibliographystyle{plainnat}
\end{document}